\documentclass[runningheads]{llncs}
\usepackage{graphicx}
\usepackage{amsmath,amssymb} 

\usepackage{algorithm}
\usepackage[noend]{algpseudocode}
\algnewcommand\algorithmicinput{\textbf{Input:}}
\algnewcommand\Input{\item[\algorithmicinput]}

\algnewcommand\algorithmicoutput{\textbf{Output:}}
\algnewcommand\Output{\item[\algorithmicoutput]}

\usepackage{color}
\usepackage{subcaption}
\usepackage{wrapfig,lipsum,booktabs}

\usepackage{capt-of}

\usepackage{multirow}

\usepackage{float}
\usepackage{dblfloatfix}
\usepackage{array}
\newcolumntype{P}[1]{>{\centering\arraybackslash}p{#1}}

\begin{document}

\title{Stacked Wasserstein Autoencoder} 

\titlerunning{Stacked Wasserstein Autoencoder}

\author{Wenju Xu \and
Shawn Keshmiri \and
Guanghui Wang}
%
\authorrunning{Wenju \it{et al.}}
%

\institute{School of Engineering, University of Kansas, Lawrence, KS, USA 66045}

\maketitle

\begin{abstract}
Approximating distributions over complicated manifolds, such as natural images, are conceptually attractive. The deep latent variable model, trained using variational autoencoders and generative adversarial networks, is now a key technique for representation learning. However, it is difficult to unify these two models for exact latent-variable inference and parallelize both reconstruction and sampling, partly due to the regularization under the latent variables, to match a simple explicit prior distribution. These approaches are prone to be oversimplified, and can only characterize a few modes of the true distribution. Based on the recently proposed Wasserstein autoencoder (WAE) with a new regularization as an optimal transport. The paper proposes a stacked Wasserstein autoencoder (SWAE) to learn a deep latent variable model. SWAE is a hierarchical model, which relaxes the optimal transport constraints at two stages. At the first stage, the SWAE flexibly learns a representation distribution, i.e., the encoded prior; and at the second stage, the encoded representation distribution is approximated with a latent variable model under the regularization encouraging the latent distribution to match the explicit prior. This model allows us to generate natural textual outputs as well as perform manipulations in the latent space to induce changes in the output space. Both quantitative and qualitative results demonstrate the superior performance of SWAE compared with the state-of-the-art approaches in terms of faithful reconstruction and generation quality.
\end{abstract}





\section{Introduction}
Recent work on deep latent variable models, such as variational
autoencoders \cite{kingma2013auto} and generative
adversarial networks \cite{goodfellow2014generative}, have
shown significant progress in learning smooth representations
of complex and high-dimensional data. These latent variable representations facilitate the
ability to apply smooth transformations in latent space in order
to produce complex modifications of the generated outputs,
while still remain on the data manifold.
Learning latent variable models is a challenging problem. Initial work on
VAEs has shown that optimization is difficult when there are
large variations in the data distribution, as the
generative model can easily degenerate with blurry reconstructions. In contrast, generative adversarial networks (GANs) \cite{goodfellow2014generative}, come without an encoder, have generated more impressive results in terms of the visual quality of images sampled from the model.

Specifically, most of the existing methods
are designed to approximate the data distribution on a single
scale. Due to the difficulty in directly approximating the high-resolution data distribution such as images, most previous methods are
limited to generating low-resolution images. To circumvent
this difficulty, we observe that real-world data, especially
natural images, can be modeled at different scales.

\begin{figure}[t!]
	\centering
	\begin{subfigure}[t]{0.7\textwidth}
		\centering
		\includegraphics[height=3.5cm]{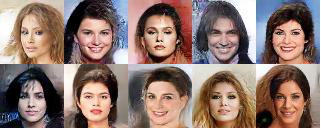}
		\caption{\label{celeba_recon_AAAE} SWAE samples with $z\sim N(0,I)$ }
	\end{subfigure}
	
	\begin{subfigure}[t]{0.7\textwidth}
		\centering
		\includegraphics[height=3.5cm]{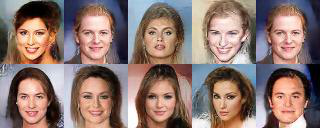}
		\caption{\label{celeba_recon_WAE} SWAE samples with $z\sim U(-1,1)$}
	\end{subfigure}
	
	\caption{\label{celeba_recon} Random samples of the proposed model trained on different prior distribution options: (a) a Gaussian prior distribution; and (b) an uniform prior distribution.}	
\end{figure}

In this work, we propose a two-stage regularized autoencoder. The proposed model is built on the theoretical analysis presented in \cite{tolstikhin2017wasserstein,kim2017adversarially}. Similar to the ARAE \cite{kim2017adversarially}, our model provides flexibility in learning an autoencoder from the input space at the first stage. The encoder is adversarially regularized to encode a continuous latent space without explicit structure. On top of this encoded prior space, we stack another autoencoder to approximate the learned prior distribution with an explicitly simple distribution, such as Gaussian.

Under this two-stage setup, this stacked Wasserstein autoencoder (SWAE) approximates the data space at two scales. It first learns a flexible autoencoder, which tends to produce faithful reconstructions of the inputs. But the encoded representation does not lay in an explicit distribution. By taking this flexibly learned representation as a prior, we can learn a latent variable model to approximate this simplified low-dimensional distribution with regularization encouraging the encoded distribution to match an explicit prior, e.g., Gaussian and uniform distribution. By combining the two models together, we are able to generate varied unseen samples given the random samples of the explicit prior, and generate consistent image manipulations by moving around in the latent space via interpolation and offset vector arithmetic. Extensive experiments demonstrate the effectiveness of our method in terms of image quality of generation and reconstruction. The main contributions of this work are listed below.
\begin{itemize}
	\item A novel latent variable model, named as the stacked Wasserstein autoencoder (SWAE), is proposed to approximate the complex and high-dimensional data distribution.
	
	\item The optimal transport is minimized at two stages. This two-step setting jointly encourages to approximate the data space while learning the encoded latent distribution as a nice explicit manifold structure. 
	
	\item We experimentally show that the SWAE model learns semantically meaningful latent variables of the observed data, enables the interpolation of the latent representation and semantic manipulation, and it can be generalized to sample unobserved data.
\end{itemize} 

The remainder of this paper is organized as follows. We describe the background of this problem and review the recent literatures in Section \ref{background}. The proposed approach is elaborated in details in Section \ref{SWAE}. Section \ref{experiment} presents both the qualitative and quantitative results and analysis. Finally, this paper is concluded in Section \ref{conclusion}.

\section{Background and Related Work}\label{background}
Deep generative models have recently received increasing attentions. They learn to approximate implicit probability distributions. Given the data sample $x\sim p_x$, where $p_x:=p(x)$ is the true while unknown distribution, and $x \in \{x_i\}_{i=1}^N$ is the observed training data, the purpose of generative model is to fit the data samples with the model parameters $\psi$ and random code $z$ sampled from an explicit prior distribution $p_z:=p(z)$. This process is denoted by $x \sim p_(x|z)$ and the training is to model a neural network $G$ that maps the representation vectors $Z$ to data $X$. 

\subsection{Regularized Autoencoder}
Unregularized autoencoders (AE) can learn an identity mapping such that the encoded latent code space can compactly capture the meaningful features to represent the observed
data. However, this latent code space is free of any structure, degenerating the capability of sampling from the latent code space. One popular approach to solve this issue is to regularize through an
explicit prior on the code space and employ a variational approximation
to the posterior, leading to a family of models called
variational autoencoders (VAE).

The VAE formulation relies on a random encoder mapping function $G$, and takes a 'reparametrization trick' to optimize the parameter. Moreover, minimizing the $KL$ divergence drives the $q_(z|x=x_i)$ to match the prior $p(z)$, thus the solution will converge close to the optima. One possible extension is to force the mixture $q_z:=\int q(z|x)dp_x $ to match the prior. With this observation, AAE \cite{makhzani2015adversarial} and WAE \cite{tolstikhin2017wasserstein} regularize the latent code space with adversarial training. WAE minimizes a relaxed optimal transport by penalizing the divergence between $q_z$ and $p_z$ as
\begin{equation*}
\begin{aligned}
D_{WAE}(P_X,P_G) &=\inf_{p(z|x)\in p_z}\mathbb{E}_{ P_x}\mathbb{E}_{p(z|x)}[c(x,G(z))] \\& + \lambda D_{z}(q_z,p_z) 
\end{aligned}
\end{equation*} 

This formulation attempts to match the encoded distribution of the training examples $p_{\theta} = \mathbb{E}_{p_x} [p(z|x)]$ to the prior $p_z$ as measured by any specified divergence
$D_{z}(q_z,p_z)$ in order to guarantee that the latent codes provided to the decoder are informative
enough to reconstruct the encoded training examples. It also allows the non-random encoders deterministically to map the inputs to their latent codes. This gives rise to the potential of unifying two types of generative models \cite{hu2017unifying,mescheder2017adversarial,larsen2016autoencoding,makhzani2017pixelgan} in one framework. There are some works on making
the prior more flexible through explicit parameterization \cite{kim2017adversarially}. In \cite{xu2019}, the authors show that standard deep architectures can adversarially approximate to the latent space and explicitly represent factors of variation for image generation.

\subsection{Generative Adversarial Network}
Deep neural network models have shown great success in many pattern recognition \cite{yu2016deep,yu2017multitask,yu2019spatial,zhou2014smart} and computer vision applications \cite{cen2019dictionary,he2018learning,ma2018mdcn,yu2018leveraging,xu2019towards}. The deep generative network is one of the most successful models for a large variant of computer vision tasks, such as high resolution image generation \cite{ledig2017photo} and image translation \cite{isola2017image,zhu2017unpaired,XU2019570}. 
The success of GANs on images have inspired many researchers
to consider applying GANs as a metric to match two distributions. To approximate the true distribution $p(x)$, the model is trained by introducing a second neural network as a discriminator 
\begin{equation*}
\begin{aligned}
D_{GAN}(p_X, p_G) &= \mathbb{E}_{x\sim p_x}[\log D(x)] \\&+ \mathbb{E}_{x\sim p_x, z\sim q_(z|x) }[\log(1-D(G(z)))]
\end{aligned}
\end{equation*}

The discriminator $D$ can provide a measure on how probable the generated sample is from the true data distribution. WGAN \cite{arjovsky2017wasserstein,gulrajani2017improved} is trained using Wasserstein-1 distance to strengthen the measure on the probability divergence and thus improves the training stability. However, the original GANs do not allow inference of the latent code. To solve this issue, BEGAN \cite{berthelot2017began} applies an auto-encoder
as the discriminator. ALI \cite{dumoulin2016adversarially} and BIGAN \cite{donahue2016adversarial} propose to
match in an augmented space by simultaneously training the model and an inverse mapping from
the random noise to the data. However, the ALI model tends to generate reconstructions that are not necessarily faithful reproductions of the inputs, the so called {\it non-identifiability} issue. To solve this problem, ALICE \cite{li2017alice} extends the ALI model to combine the framework of cross entropy (CE). This additional regularization imposes a restriction on the connection between the image and the latent variable, and thus enables the faithful reconstruction. A recent successful extension, VEEGAN \cite{srivastava2017veegan}, is also trained by discriminating the joint samples of the data and the corresponding latent variable $z$, by introducing an additional regularization to penalize the cross entropy of the inferred latent code.

\subsection{Stacked Model}
A number of works have been proposed to use multiple GANs to improve sample quality. LAPGANs \cite{denton2015deep} is built on a series of GANs within a
Laplacian pyramid framework. For each generator, the StackGANs \cite{huang2017stacked,zhang2017stackgan,zhang2017stackgan2}  generate high-resolution images that are conditioned on their low-resolution inputs. At the $i_{th}$ level, a discriminator $D_i$ is trained to distinguish the generated representations $G_i(h_{i+1}, z_i)$ from encoded 'real' representations $h_i$.
\begin{equation*}
\begin{aligned}
Di &= \mathbb{E}_{h_i\sim P_{data,E}} [log D_i(hi)] \\&+ \mathbb{E}_{z_i\sim P_{zi}, h_{i+1}\sim P_{data,E}} [log(1 - D_i(G_i(h_{i+1}, z_i)))] 
\end{aligned}
\end{equation*}
where $h_i$ and $h_{i+1}$ are the encoded representations, and $z$ is the random noise. Our proposed model differs from existing regularized autoencoder models in that it learns a hierarchical latent space, and only matches the encoded latent distribution to explicit prior at the second stage. 

\begin{figure}[!h]
	\begin{center}		
		\includegraphics[width=0.52\textwidth]{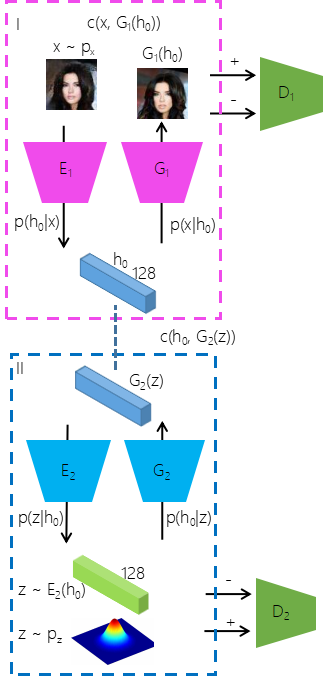}
	\end{center}
	\caption{An overview of the SWAE architecture. Red rectangle contains the stage-I model, which learns a flexible model to provide 'real' encoded representation $h_0$. Blue rectangle contains the stage-II model that learns a latent variable model regularized to match the latent space to an explicit prior distribution.}
	\label{f_AAAE}
\end{figure}

\section{Proposed Method}\label{SWAE}
To build an autoencoder for faithful reconstruction with a nice latent manifold structure, we propose to learn stacked autoencoders at two stages, as shown in Figure \ref{f_AAAE}. The proposed SWAE consists of two major components: The encoder-generator, $E_1, G_1$, at the first stage and the second encoder-generator, $E_2, G_2$, at the second stage. At each stage, we adversarially train the encoder-generator with additional discriminators $D_1, D_2$. In this work, we aim at minimizing optimal transport $Wc(P_X, P_G)$ at two scales. Given the true (but unknown) data
distribution $P_X$, at the first stage, it learns a latent variable model $P_G$  specified by the encoded prior distribution $P_{h_0}$ of the latent codes and the generative model $G_1(h_0)$ of the data points $x\in X$ given $h_0$. We assume that the successfully trained autoencoder ensures $p(h_0|x)$, the output of $E_1$, is the true latent codes distribution with an unknown structure, which cannot be sampled in a closed form. To solve the sampling issue, at the second stage, we train another encoder-generator by minimizing optimal transport $Wc(H, P_{G_2})$ between the encoded (but unknown) latent variable distribution $P(h_0|x)$ and a latent variable model $G_2(z)$ of the latent encoder prior $h_0\in H$ given $z$. And $p(z|h_0)$, the output of $E_2$, is enforced to match the explicit prior $p_Z$. The joint objective is defined as 
\begin{align}
\begin{aligned}
\mathcal{O}_{SWAE}(&E_1, E_2, G_1, G_2, D_1, D_2):= W_c(P_X, P_G)  \\ 
&:=\inf_{p(h_0|x)\in H} \mathbb{E}_{ P_x}\mathbb{E}_{p(h_0|x)}[c(x,G_{1}(h_0))] \\&+ \inf_{p(z|h_0)\in P_Z} \mathbb{E}_{ P_{h_0}}\mathbb{E}_{p(z|h_0)}[c(h_0,G_{2}(z))] \\&+ D_z(P_Z, P_{E_2})   
\end{aligned}
\label{objective}
\end{align}
where $P_{E_2}$ is the output distribution of encoder $E_2$ and $D_z$ is an arbitrary divergence between
$P_Z$ and $P_{E_2}$.

The above objective is not easy to solve. We attempt to optimize each term by considering: (1)
the first encoder-generator to minimize data reconstruction; (2) the
second encoder-generator to learn a latent variable model; and (3) the encoder-generator adversarially to minimize $W_c$. In the following, we discuss how to simplify and transform the cost function into a computable version at each stage.
\begin{figure}[t]
	\begin{subfigure}[t]{0.5\textwidth}
		\centering
		\includegraphics[height=6cm]{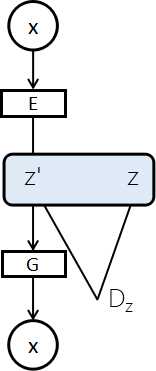}
		\caption{\label{fig:cifar} WAE model}
	\end{subfigure}
	\begin{subfigure}[t]{0.5\textwidth}
		\centering
		\includegraphics[height=6cm]{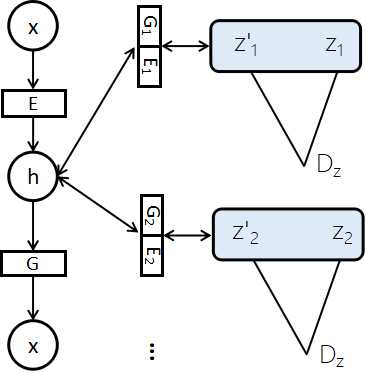}
		\caption{\label{fig:cifar} Our model}
	\end{subfigure}
	\caption{\label{prior}Schematic comparison of WAE and the proposed SWAE models, where $D_z$ is a divergence between encoded $z'$ and $z$, a sample of prior distribution. Due to the two-step setup, multiple stage-II models can be trained at the same time.}	
\end{figure}

\begin{algorithm}
	\caption{The training pipeline of SWAE.
	}\label{algo}
	\begin{algorithmic}[1]
		\Input Source training images; \\
		\hskip\algorithmicindent Initialize the parameters of the encoder $E_1$, $E_2$, \\
		\hskip\algorithmicindent generator $G_1$, $G_2$, the discriminator $D1$, $D_2$. \\ \hskip\algorithmicindent Regularization coefficient $\lambda > 0$;
		\For {$i\in \{1...N\}$}
		\State Sample $x^i \sim p(x)$; \\
		\hskip\algorithmicindent  Sample $h_0^i \sim p(h_0|x)$ as $h_0^i = E_1(x^i)$.
		
		\State
		\State Update $D_1$ by ascending:\\		\vspace*{1mm}
		\hskip\algorithmicindent \hskip\algorithmicindent $ \frac{\lambda}{n} \sum_{1}^{n}\log D_1(x^i) + \log(1-D_1(G_1(h_0^i)))$
		\vspace*{1mm}
		\State Update $E_1$ and $G_1$ by ascending: \\
		\vspace*{1mm}
		\hskip\algorithmicindent \hskip\algorithmicindent $
		\frac{\lambda}{n} \sum_{1}^{n}||x^i, G_1(h_0^i)||_2^2 - \lambda \log D_1(G_1(h_0^i)))$	
		\vspace*{.4mm}
		\State 	
		\For {$j\in \{1...k\}$}	
		\State Sample $h_0^j \sim p(h_0|x^j)$; \\ \hskip\algorithmicindent \hskip\algorithmicindent Sample $z^j \sim p_Z$.
		\State				
		\State Update $D_2$ by ascending:\\		\vspace*{1mm}
		\hskip\algorithmicindent \hskip\algorithmicindent $  \hskip\algorithmicindent \frac{\lambda}{n} \sum_{1}^{n}\log D_2(z^j) + \log(1-D_2(E_2(h_0^j)))$
		\vspace*{1mm}
		\State Update $E_2$ and $G_2$ by ascending: \\
		\vspace*{1mm}
		\hskip\algorithmicindent \hskip\algorithmicindent $  \hskip\algorithmicindent
		\frac{\lambda}{n} \sum_{1}^{n}||h_0^j, G_2(z^j)||_2^2 - \log D_2(E_2(h_0^j))$
		\EndFor
		\EndFor	 	
	\end{algorithmic}
\end{algorithm}
\textbf{Stage I:}
Instead of enforcing the encoded latent distribution to match an explicit prior, we simplify the task by first learning a flexible latent variable model, which aims at faithful reconstruction for observed data. As a result, the encoded latent space exactly reflects the data variation. Stage-I SWAE consists of the encoder $E_1$ and generator $G_1$. They are adversarially trained with discriminator $D_1$  by maximizing
\begin{align}
\min_{E_1, G_1} \max_{D_1} \mathbb{E}_{ P_x}\mathbb{E}_{p(h_0|x)}[c(x,G_{1}(h_0))]
\end{align}
for the measurable cost function $c((x,G_{1}(h_0)))$, we use a squared cost function and a weighted adversarial objective
\begin{align}
\begin{aligned}
c(x,G_{\psi})  &= ||x - y||^2_2 + \lambda D_{GAN}(x, G_{1}(h_0) \\
&= ||x - y||^2_2 + \lambda[\log D_1(x)] \\&+ [\log(1-D_1(G_1(h_0)))])
\end{aligned}
\end{align}
where $D_{GAN}$ is the adversarial loss between $x$, the sample of data distribution, and $G_{1}(h_0)$, the output of generator model $G_1(h_0)$. Since this autoencoder is trained without direct regularization under the latent space, the adversarial training process is free of model collapse and assists to generate sharp image samples.

\textbf{Stage II:}
The flexibly encoded representation $h_0$ from Stage-I could be considered as a 'real' sample of the true distribution $H$, but it is free of any explicit structure. It is difficult to sample directly for $x\sim p(x|h_0)$. The Stage II model is to approximate the encoded representation space $H$ with a latent variable model specified by an explicit simple prior distribution. The Stage-II consists of the encoder $E_2$ and generator $G_2$. The discriminator $D_2$ is employed to enforce the match between $P_Z$ and $P_{E_2}$. The objective function is defined as
\begin{align}
\begin{aligned}
\min_{E_2, G_2} \max_{D_2} &\mathbb{E}_{ P_{h_0}}\mathbb{E}_{p(z|h_0)}[c(h_0,G_{2}(z))] \\&+ D_z(P_Z, P_{E_2})
\end{aligned}
\end{align}

This objective could be consider as minimizing the optimal transport $Wc(H, P_{G_2})$ between the encoded (but unknown) representation distribution $H$ and the output distribution of the latent variable model $P_{G_2}$. Here, we use the same squared cost function but without adversarial objective.
\begin{align}
\begin{aligned}
c(h_0,G_{2}(z))  = ||h_0 - G_{2}(z)||^2_2 
\end{aligned}
\end{align}
Specifically, we introduce an adversary (discriminator $D_2$) in the
latent space $Z$ trying to separate the ``true" points sampled from $P_Z$ and the ``fake" ones sampled from $P_{E_2}$.
\begin{align}
\begin{aligned}
D_z(P_Z, P_{E_2}) 
&=  \mathbb{E}_{p_z}[\log D_2(z)] \\&+ \mathbb{E}_{p(h_0|x) }[\log(1-D_2(E_2(h_0)))]
\end{aligned}
\end{align}

The full training process is outlined in Algorithm \ref{algo}.

\subsection{Stacked GAN-based $D_z$}
Empirically the choice of the prior distribution
$P_Z$ strongly influences the performance of the generative models. The simplest choice is to employ a fixed distribution such as Gaussian distribution. However, this choice is seemingly too constrained to achieve a faithful reconstruction and even suffers from mode collapse. Our model exploits the two-stage setup, which first map the complex, high-dimensional data distribution to a low-dimensional representation, and then learn a latent variable model to approximate the representation distribution. Therefore, it is not sensitive to the choice of a prior distribution and we can stack several encoder-generators to learn multiple latent variable models as illustrated in Figure \ref{prior}. The trained model enables us to draw samples given different prior distributions. 

\subsection{Connection to WAE}
The optimal
transport (OT) problem in \cite{tolstikhin2017wasserstein} is defined as:
\begin{equation*}
\begin{aligned}
Wc(P_X, P_G) := \inf_{\Gamma\in(X\sim P_X,Y \sim P_G)}\mathbb{E}_{(X,Y)\in \Gamma}[c(X, Y )]
\end{aligned}
\end{equation*} 

The WAE proves that
learning an autoencoder can
be interpreted as learning a generative model with latent variables,
as long as we ensure that the marginalized encoded
space is the same as the prior.
\begin{equation*}
\begin{aligned}
W_{c}(P_X, P_G) &:=\inf_{Q: Q_z=P_z}\mathbb{E}_{ P_x}\mathbb{E}_{Q(Z|X)}[c(X,G(Z))]
\end{aligned}
\end{equation*} 

In practice, learning the marginalized encoded space to be the same as the prior is nontrivial. Thus, we seek to approximate the prior distribution at two stages:
\begin{equation*}
\begin{aligned}
W_{c}(P_X, P_G) &:=\inf_{Q: Q_h=P_h}\mathbb{E}_{ P_x}\mathbb{E}_{Q(H|X)}[c(X,G(H))] \\&+ \inf_{Q: Q_z=P_z}\mathbb{E}_{ P_x}\mathbb{E}_{Q(Z|H)}[c(H,G_h(Z))]
\end{aligned}
\end{equation*} 
where the stage-I model aims at generating the representation distribution $Q_h$ by minimizing the first term, while the second term is to learn a latent variable model specified by an explicit prior $P_z$. 

\section{Experiments} \label{experiment}
In this section, we conduct extensive experiments to evaluate the proposed SWAE model. Three publically available datasets are used to train the model: MNIST consisting of 70k images, CIFAR-10 \cite{krizhevsky2009learning} consisting of 60k images in 10 classes, and CelebA \cite{liu2015deep} containing roughly 203k images. The performances of our approach are quantitatively and qualitatively compared with the state-of-the-art approaches. We report our results on three aspects of the model. First, we measure the reconstruction accuracy of the observed data inputs and the quality of the randomly generated samples. Next, we explore the latent space by manipulating the codes for consistent image transformation \cite{kingma2018glow,larsen2015autoencoding}. Finally, we study the crucial aspect that affects the performance of both the reconstruction and random generation.
\begin{figure}[t]
	\begin{subfigure}[t]{0.5\textwidth}
		\centering
		\includegraphics[height=5cm]{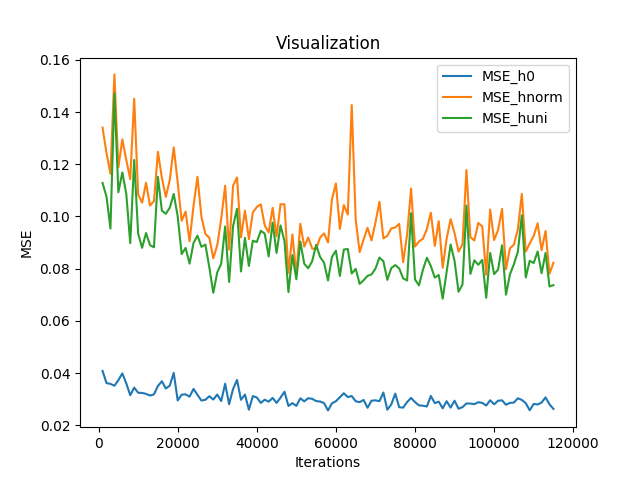}
	\end{subfigure}
	\begin{subfigure}[t]{0.5\textwidth}
		\centering
		\includegraphics[height=5cm]{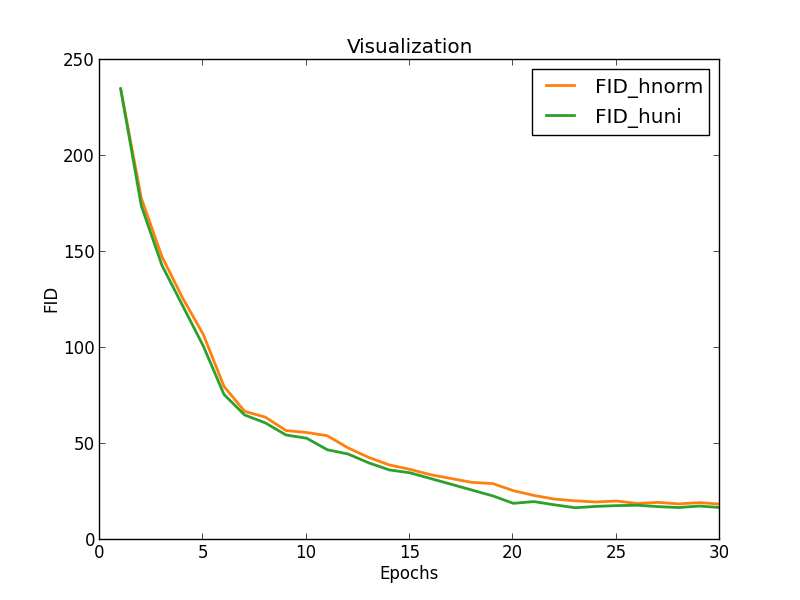}
	\end{subfigure}
	\caption{\label{training} Illustration of the training process on CelebA. Left: mean squared errors (MSE) of the input images and the reconstructions conditioned on different latent codes. Right: the FID scores of random generations after each training epoch.
	}		
\end{figure}

\textbf{Experiment setup:}
All models were optimized via Adam \cite{kingma2014adam} with a learning rate of 0.0001. We set $\lambda = 0.001$, and $k = 2$. We do not perform any dataset-specific tuning except for employing early stopping based on the average data reconstruction loss of $x$ on the validation sets. For the CelebA dataset, we crop the original images from $178\times218$ size to $178\times178$ centered at the faces, then resize them to $64\times64$. The training process is shown in Figure \ref{training}. The MSE of reconstructions from $h_0$ is constantly low, which means that the stage-I model easily encodes the representation distribution and is reasonable to provide 'real' samples for training the stage-II model. A smooth learning process at the stage-I will provide constant 'real' representation. We prefer to set a big batch size with a value of 64, which is significant for training stabilization. We adopt the patch discriminator \cite{isola2017image,zhu2017unpaired} for $D_1$. The architectures for the model are provided in the supplemental material.

\noindent\textbf{Quantitative evaluation protocol:} We adopt the mean squared error (MSE) and the inception score (ICP) \cite{salimans2016improved,li2017alice} to quantitatively evaluate the performance of the generative models. MSE is employed to evaluate the reconstruction quality, while ICP reflects the plausibility and variety of the sample generation. Based on the pretrained inception model C, the ICP score is calculated by
\begin{equation*}
\begin{aligned}
ICP~score = exp(\mathbb{E}_{x\in X} [KL(C(x)||C(G(z)))])
\end{aligned}
\end{equation*}  
where $KL$ denotes the Kullback$-$Leibler divergence and a higher ICP score indicates better performance. In order to quantitatively assess the quality of the generated images on the CelebA dataset, we adopt the Frechet inception distance (FID) introduced in \cite{heusel2017gans}. The FID score measures the distance between the Gaussian distribution with mean and covariance $(m, v)$ of the real data and the Gaussion distribution $(m_{\omega}, v_{\omega})$ of the generated data. It is calculated by
\begin{equation*}
\begin{aligned}
FID~score = ||m-m_{\omega}||_2^2 + Tr(v+v_{\omega} -2(v\times v_{\omega})^{1/2})
\end{aligned}
\end{equation*} 

In our experiments, the ICP and FID scores are computed statistically based on $10,000$ samples.

\begin{table}
	\centering
	{
		\renewcommand{\arraystretch}{1.3}
		\centering
		\caption{Quantitative results on real-world datasets. To compare the quality of random samples, we report ICP scores (higher is better) on MNIST, CIFAR-10 data, and FID scores (smaller is better) on CelebA. For the reconstruction quality, we report MSE (smaller is better). $\dag$ is the best performance reported in \cite{li2017alice}; $\ddag$ is calculated using the method in \cite{li2017alice}. Comparing to the value in \cite{tolstikhin2017wasserstein}, $\star$ is degraded due to a different crop style.}
		\label{t1}
		\begin{tabular}{P{3.4cm}|P{1cm}P{1cm}|P{1cm}P{1cm}|P{1cm}P{1cm}}
			
			\hline\hline
			\multirow{2}{*}{Settings} & \multicolumn{2}{|c|}{MNIST}&\multicolumn{2}{|c|}{CIFAR-10} &\multicolumn{2}{|c}{ CelebA} \\ 
			\cline{2-7}
			& ICP&MSE& ICP&MSE& FID&MSE\\ 
			\hline
			True data  & &&& &1.94&\\ 
			\hline\hline
			ALI &  $8.84^{\dag}$ &$0.38^{\dag}$&$5.97^{\dag}$ &$0.560^{\dag}$&6.95&0.281\\ 
			ALICE & $9.35^{\dag}$&$0.07^{\dag}$&$6.04^{\dag}$ &$0.214^{\dag}$ &&\\
			WAE &    &&&&$98.78^{\star}$&0.020\\ 
			SWAE (w/o $D_1)$ &  &&& &68.15&0.046\\ 
			SWAE (norm)&$8.87^{\ddag}$  &0.01&5.73&0.081 &18.38&0.072\\ 
			SWAE (unif)&$8.91^{\ddag}$  &0.01&5.81&0.078 &17.14&0.066\\			
			\hline		
		\end{tabular}
	}
\end{table}
\begin{figure}[h!]
	\begin{subfigure}[t]{0.5\textwidth}
		\centering
		\includegraphics[height=4.0cm]{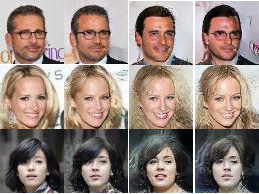}
		\caption{\label{swae_rec} Reconstructions of SWAE conditioned on different latent codes. From left to right: input, $G_1(h_0)$, $G_1(h_{norm})$, $G_1(h_{uni})$}
	\end{subfigure}
	\hspace{5mm}
	\begin{subfigure}[t]{0.5\textwidth}
		\includegraphics[height=4.0cm]{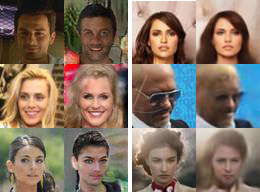}
		\caption{\label{ali_wae_rec} Reconstructions of ALI (left) and WAE (right),  where the odd columns are inputs and even columns are reconstructions. }
	\end{subfigure}
	
	\begin{subfigure}[t]{0.5\textwidth}
		\centering
		\includegraphics[height=4.0cm]{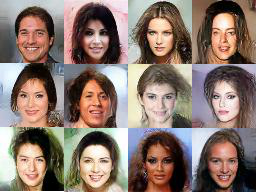}
		\caption{\label{sample_uni}SWAE samples with $z\sim U(-1,1)$}
	\end{subfigure}
	\begin{subfigure}[t]{0.5\textwidth}
		\centering
		\includegraphics[height=4.0cm]{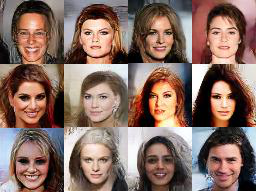}
		\caption{\label{sample_lat}SWAE samples with $z\sim N(0,I)$}
	\end{subfigure}
	
	\begin{subfigure}[t]{0.5\textwidth}
		\centering
		\includegraphics[height=4.0cm]{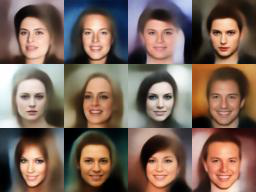}
		\caption{\label{sample_wo}SWAE samples w/o $D_1$}
	\end{subfigure}	
	\hspace{5mm}		
	\begin{subfigure}[t]{0.5\textwidth}
		\includegraphics[height=4.2cm]{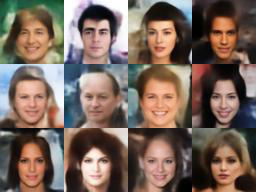}
		\caption{\label{wae}WAE samples}
	\end{subfigure}
	\caption{\label{sam_rec} Comparison of the reconstructions and generations on the CelebA dataset.}	
\end{figure}

\subsection{Random Samples and Reconstruction}
The proposed method maps input data to two types of latent codes. At the first stage, we learn a flexible encoded distribution $H$, which tightly captures useful features that represent the observed inputs. At the second stage, the latent space distribution $P_{E_2}$ is regularized to match an explicit prior. Here, we learn two latent variable models specified by Gaussian and uniform distribution, respectively. We begin our experiments by comparing our model against two closely related state-of-the-art approaches: ALI \cite{dumoulin2016adversarially} and WAE \cite{tolstikhin2017wasserstein}. The quantitative results are tabulated in Table \ref{t1}. SWAE is able to generate impressive synthesized images, achieving MSE (0.01) and ICP (8.91) on MNIST. This outperforms GAN based model, ALI (MSE 0.38 and ICP 8.84), while also being competitive to the modified ALICE (MSE 0.07 and ICP 9.35). As for more complicate datasets, such as CelebA, the ALI generates high quality samples (FID 6.95), however, it fails to faithfully reconstruct the input images. This is evidenced by the high reconstruction err (MSE 0.281). While the proposed SWAE dose not have this issue. It achieves better performance in terms of both FID (17.14) and MSE (0.066). This is due to the stacked structure of our model. At the first stage, it prefers to high-quality reconstruction; and it learns a latent variable model for random generation at the second stage. This two-steps learning scheme enables our model to work well in both random generation and faithful reconstruction.

\begin{figure}[h!]
	\begin{center}		
		\includegraphics[height=5.5cm]{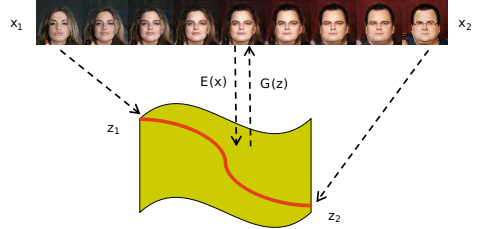}
	\end{center}
	\caption{Illustration of the manifold. By manipulating the latent variables encoded from two images, it is able to generate interpolations between these two inputs.}
	\label{manifold}
\end{figure}

Figure \ref{swae_rec} and Figure \ref{ali_wae_rec} show some comparative reconstructions by SWAE, ALI, and WAE, respectively. It is evident that the reconstructions of ALI are not faithful reproduction of the input data, although they are related to the input images. The results demonstrate the limitation of adversarial regularization in reconstruction. This is also consistent with the results in terms of MSE as shown in Table \ref{t1}. Some generated samples are shown in Figure \ref{sample_uni} and \ref{sample_lat}. We observe that the $D_1$ is crucial for the image quality. Figure \ref{sample_wo} shows the blurry random generations of SWAE model trained without $D_1$, ($\lambda = 0$). The quantitative FID score (68.15) also reflects the degradation of image quality. However, when $\lambda \geq 0.01$, we observe serious artifacts in the generated samples. Similarly, the original WAE dose not integrate this discriminator and could only generate blurry samples (FID 98.78). The random samples are shown in Figure \ref{wae}. It is clear that the adversarial learning assists to generate sharp images matching the true data distribution.

The number of iteration of the inner loop in Algorithm \ref{algo} will also affect the performance. This inner loop is to update the latent variable model to approximate the encoded representation distribution. The latent variable model is not able to follow the change of the encoded representation distribution when $k=1$. We also observe that setting $k > 2$ works well for the latent variable model to approximate the encoded representation distribution. As a trade-off of efficiency and effectiveness, we prefer to set $k=2$.

\begin{figure}[ht]
	\begin{center}		
		\includegraphics[width=1\textwidth]{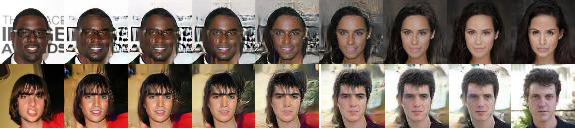}
	\end{center}
	\caption{Latent space interpolations. The leftmost and rightmost columns are the original data pairs, while the columns in between are the reconstructions generated with the linearly interpolated latent codes.}
	\label{interp}
\end{figure}
\begin{figure}[ht!]
	\begin{center}		
		\includegraphics[width=1\textwidth]{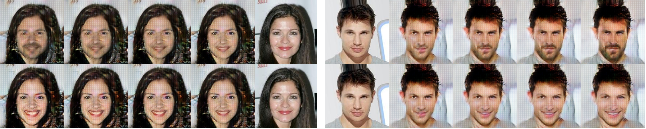}
	\end{center}
	\caption{Manipulation of attributes of a face. Each row is made by interpolating the latent code of an
		image along a vector corresponding to the attribute, with the middle image being the original image. First row: to mustache; second row: to smile.}
	\label{interp_att}
\end{figure}
\begin{figure}[h!]
	\begin{center}		
		\includegraphics[width=1\textwidth]{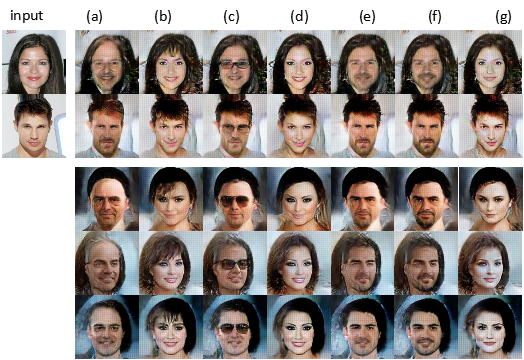}
	\end{center}
	\caption{Using SWAE to reconstruct samples with manipulated latent codes. (a) bald; (b) bangs; (c) eyeglasses; (d) heavy makeup; (e) male; (f) mustache; (g) pale skin. The manipulations in the first two rows are conditioned on input images; and the results in the last three rows are random generations given the attributions.}
	\label{semantic}
\end{figure}
\begin{figure}[h!]
	\centering
	\begin{subfigure}[t]{1\textwidth}
		\centering
		\includegraphics[height=4.5cm]{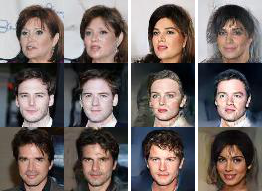}
		\caption{\label{celeba_recon_ALI}SWAE reconstructions conditioned on different latent codes. From left to right: input, $G_1(h_0)$, $G_1(h_{norm})$, $G_1(h_{uni})$}
	\vspace{2mm}
	\end{subfigure}
	\vspace{2mm}
	\begin{subfigure}[t]{0.48\textwidth}
		\centering
		\includegraphics[height=4.5cm]{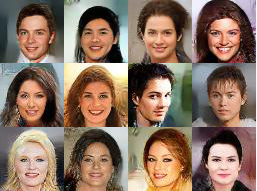}
		\caption{\label{celeba_recon_ALI}SWAE samples with $z\sim N(0,I)$}
	\end{subfigure}
	\vspace{2mm}
	\begin{subfigure}[t]{0.48\textwidth}
		\centering
		\includegraphics[height=4.5cm]{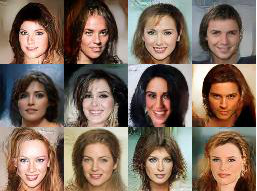}
		\caption{\label{celeba_recon_ALI}SWAE samples with $z\sim U(-1,1)$}	
	\end{subfigure}
	
	\begin{subfigure}[t]{1\textwidth}
		\centering
		\includegraphics[height=1.65cm]{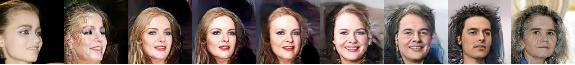}
		\caption{\label{slide_bad}Latent space interpolations}	
	\end{subfigure}
	\caption{\label{epoch} Performance of the model trained after 50 epochs. The sample quality is improved, while the reconstructions are not faithful to the inputs and therefore the latent manifold structure disappears.}
\end{figure}

\subsection{Latent Space Interpolation}
The latent variable model is characterized by learning semantic representations of the observed data. The latent variables are disentangled and evenly distributed in a well-organized manifold structure. Figure \ref{manifold} demonstrates
the learned manifold. To explore the latent manifold structure \cite{yin2018locally,zhang2019nonlinear}, we investigate the latent space interpolations between the example pair $(x_1,x_2)$ by linearly interpolating between $h_{x_1} = E_1(x_1)$ and $h_{x_2}=E_1(x_2)$ with equal steps in the latent space. We observe smooth transitions between the pairs of examples, and intermediate images remain plausible and realistic as shown in Figure \ref{interp}. Figure \ref{interp_att} illustrates the interpolations between images with two different attributes.

\subsection{Semantic Manipulation}
Learning disentangled latent features is an important computer vision topic. It learns the latent codes to represent different attributes of the observations \cite{tran2017disentangled,zhang2017age,donahue2016adversarial}. To demonstrate the capability of learning disentangled latent codes, we cluster the learned latent codes according to image attributes. We then calculate the average latent vector $h_{pos}$ for images with the attribute and $h_{neg}$ for images without, and then use the difference $(h_{pos} - h_{neg})$ as a direction for manipulating. This is done after the model is trained making it extremely easy to perform for a variety of different target attributes. 
\begin{equation*}
\begin{aligned}
x^{\prime} = G_1(h + \lambda_h(h_{pos} - h_{neg}))
\end{aligned}
\end{equation*}  
where $h = G_2(z)$ given $z\sim P_Z$ or $h= E_1(x)$ given $x\sim P_X$, and $\lambda_h$ is the scale used to emphasize the added attribute. The reconstructed results are shown in Figure \ref{semantic}. It proves that SWAE can achieve reliable geometry of latent space without any class information at the training stage.

\subsection{Effect of Training Epoch}
Figure \ref{epoch} shows the random generation and reconstruction of the model trained after 50 epochs in contrast to the best model trained after 25 epochs. The model achieves a better visual quality of random generations. The FID score goes down to 14.6. However, the MSE is 0.15. At this point, the discriminator $D_1$ is more sensitive to samples not in the true data distribution. The improvement of random generation is at the cost of faithful reconstruction. The latent manifold structure is destroyed as showed in Figure \ref{slide_bad}. Thus we choose to stop the training process early as a trade-off of generation and reconstruction.

\section{Conclusion}\label{conclusion}
In this paper, we have presented a stacked Wasserstein autoencoder, which learns the latent code space a manifold structure and generates high-quality samples. The model is fulfilled by training an autoencoder in two stages with more flexibility. The first stage learns a flexible autoencoder, which tends to produce faithful reconstructions of the inputs. However, the encoded representation distribution is not an explicit distribution. With a latent variable model, the flexibly encoded representation distribution is further approximated. Experimental results demonstrate that the images sampled from the learned distribution are of better quality while the reconstructions are consistent with the inputs. 

\section*{Acknowledgement}
The work was supported in part by USDA NIFA under the grant no. 2019-67021-28996, the Research Grant Opportunity program of the University of Kansas, and the Nvidia GPU grant.\\[24pt]
\clearpage
\bibliographystyle{splncs04}
\bibliography{SWAE_arXiv}
\end{document}